\pdfoutput=1

\documentclass[11pt]{article}

\usepackage{acl}

\usepackage{times}
\usepackage{latexsym}
\usepackage{booktabs}

\usepackage[T1]{fontenc}

\usepackage[utf8]{inputenc}

\usepackage{microtype}

\usepackage{inconsolata}

\usepackage{graphicx}
\usepackage{multirow}
\usepackage{amsmath}
\usepackage{paralist}

\title{A Linguistically Motivated Analysis of Intonational Phrasing in Text-to-Speech Systems: Revealing Gaps in Syntactic Sensitivity}

\author{
Charlotte Pouw\textsuperscript{1}, 
Afra Alishahi\textsuperscript{2}, 
Willem Zuidema\textsuperscript{1} \\
\textsuperscript{1}Institute for Logic, Language and Computation, University of Amsterdam \\
\textsuperscript{2}Cognitive Science and Artificial Intelligence, Tilburg University \\
\texttt{\{c.m.pouw,w.h.zuidema\}@uva.nl} \\
\texttt{a.alishahi@tilburguniversity.edu}
}

\begin{document}

\maketitle
\begin{abstract}

We analyze the syntactic sensitivity of Text-to-Speech (TTS) systems using methods inspired by psycholinguistic research. Specifically, we focus on the generation of intonational phrase boundaries, which can often be predicted by identifying syntactic boundaries within a sentence. We find that TTS systems struggle to accurately generate intonational phrase boundaries in sentences where syntactic boundaries are ambiguous (e.g., garden path sentences or sentences with attachment ambiguity). In these cases, systems need superficial cues such as commas to place boundaries at the correct positions. In contrast, for sentences with simpler syntactic structures, we find that systems \textit{do} incorporate syntactic cues beyond surface markers. Finally, we finetune models on sentences without commas at the syntactic boundary positions, encouraging them to focus on more subtle linguistic cues. Our findings indicate that this leads to more distinct intonation patterns that better reflect the underlying structure.

\end{abstract}


\section{Introduction}

Humans use prosody to convey meaning beyond words. Intonation, an important aspect of prosody, organizes speech into meaningful units called \textit{intonational phrases} \cite{bolinger1989intonation}. Linguistic theory suggests that in human speech, the positioning of boundaries between these phrases is closely linked to syntactic structure. Some theorists claim that intonational phrasing can directly be derived from syntactic structure (e.g., \citealp{c154a936-5e4c-3719-91d1-88cf9baf71c2,cooper1980syntax}); others argue that the mapping is more complex and there must exist an independent level of intonational structure (e.g., \citealp{pierrehumbert1980phonology,selkirk1984phonology,nespor2007prosodic}). 

Regardless of the theoretical perspective, it is well-established that intonational and syntactic boundaries often overlap. Acoustic markers of intonational boundaries (i.e., pauses, syllable lengthening, and pitch contour changes) are frequently observed at syntactic boundary positions \cite{KLATT1975129, COOPER1976151, ferreira1993creation, CROFT95, watson2004relationship}. Psycholinguistic experiments have also shown that the placement of intonational boundaries influences parsing decisions in speech processing (e.g., \citealp{pynte1996prosodic,KJELGAARD1999153,10.1162/jocn.2011.21610}), and that speakers adjust their intonation to signal the underlying structure of an ambiguous sentence (e.g., \citealp{SNEDEKER2003103,KRALJIC2005194,schafer2005approaches}).

In this paper, we analyze if we can observe a similar link between syntax and intonational phrasing in the behavior of Text-to-Speech (TTS) systems. Such systems have become increasingly capable of mimicking human intonation patterns, but it remains an open question to what extent these patterns are shaped by linguistic structure. We propose to use methods from psycholinguistics to investigate this question, an approach previously used to assess
the syntactic sensitivity of text-based language models (e.g., \citealp{linzen2016assessing,futrell-etal-2019-neural,ettinger-2020-bert,jumelet2024language}). This involves the use of controlled stimuli that require a reliance on specific (linguistic) information to elicit specific behavioral responses.

We find that TTS systems incorporate syntactic information when it reliably signals the need for an intonational boundary (i.e., obvious clause boundaries in simple sentence structures), although the duration of intonational boundaries is also modulated by lexical cues. In more complex cases such as garden path sentences and attachment ambiguities, systems need explicit punctuation cues to place intonational boundaries at the correct syntactic positions. In the absence of such cues, TTS systems tend to default to the statistically most likely intonation pattern, which may not align with the underlying structure.

Encouragingly, we also find that with increased exposure to sentences where we have removed explicit punctuation cues at the intonational boundary positions, TTS systems \textit{can}, to some extent and under some conditions, learn to generate more distinct intonation patterns that better reflect alternative syntactic structures. We hope that these findings contribute to the development of more linguistically informed TTS training and evaluation paradigms. All code is available at our \href{https://github.com/CharlottePouw/interpret-tts}{GitHub repository}.

\section{Psycholinguistic Background}

The relationship between intonation and syntax has been explored in various psycholinguistic studies. These studies often use sentences with (temporary) syntactic ambiguity \cite{cutler1997prosody}, as listeners have to make a decision about the syntactic structure based on controlled evidence (e.g., the position of an intonational boundary). These sentences therefore provide a unique opportunity to study the interplay between intonational boundary placement and syntactic parsing decisions in speech processing.

A key area of research has focused on \textbf{garden path} sentences—structures that initially lead the listener to a syntactic interpretation that must later be revised \cite{bever1970cognitive}. From the extensive literature on the human processing of such sentences, we mention \citet{KJELGAARD1999153}, who examined sentences such as \textit{When Roger left the house was dark}, which initially confuses the listener into interpreting \textit{left the house} as a single constituent. They found that an intonational boundary after \textit{left} facilitated processing speed, as it helped to clarify the syntactic structure. However, a boundary after \textit{the house} led to processing difficulty because it interfered with the underlying structure.

A related phenomenon occurs with sentences that exhibit \textbf{attachment ambiguity}, where there are two alternative syntactic structures based on the attachment site of a prepositional phrase. Many psycholinguistic studies have revealed details of how humans deal with such ambiguity. For instance, 
\citet{pynte1996prosodic} showed that, in sentences such as \textit{The spies inform the guards of the conspiracy}, an intonational boundary after \textit{inform} leads to the NP-attachment interpretation (i.e., \textit{of the conspiracy} attaches to \textit{the guards}), whereas a second boundary after \textit{guards} leads to the VP-attachment interpretation (i.e., \textit{of the conspiracy} attaches to \textit{inform}). These findings illustrate how the position of intonational boundaries can guide listeners towards alternative syntactic structures.

In speech production, it has been shown that speakers adjust their intonation to signal the underlying structure of an ambiguous sentence. For example, \citet{SNEDEKER2003103} studied the placement of intonational boundaries in a referential game setting. Speakers had to refer to objects with instructions such as \textit{Tap the frog with the flower}. The attachment site of the PP \textit{with the flower} was ambiguous, as the room contained a frog toy with a flower on its head, as well as a frog and a flower separately. When speakers were aware of the ambiguity, they produced a boundary after \textit{frog} to signal the VP-attachment structure (i.e., when they wanted the addressee to use the flower as an instrument); they did not do this for the NP-attachment scenario (i.e., when they wanted the addressee to tap the frog which had the flower on its head). In other (similar) studies, this pattern has been observed even for speakers who were unaware of the potential ambiguity \cite{KRALJIC2005194, schafer2005approaches}.

Taken together, these studies illustrate how both listeners and speakers use intonational boundaries to interpret and signal syntactic structures. In the present study, we systematically analyze whether and how TTS systems are informed by syntax to determine the placement of intonational boundaries.

\section{Text-to-Speech Models}

We select three publicly available TTS systems with diverse architectures. We also provide Mean Opinion Scores (MOS) (i.e., human ratings of the naturalness of each system's output speech, on a scale from 1-5) reported for each system, while noting that these scores were not consistently measured, and should therefore only been seen as approximate \cite{kirkland2023stuck,chiang2023we,le2024limits}.

\textbf{Tacotron2} \cite{shen2018natural} is an LSTM-based encoder-decoder. The bidirectional encoder converts a character sequence into a hidden feature representation, which the decoder (with attention) takes as input to autoregressively predict spectrogram frames. A WaveNet vocoder \cite{van2016wavenet} transforms these spectrogram frames into a waveform. The model was trained on an internal US-English dataset containing 24.6 hours of speech from one female speaker. MOS: 3.52\footnote{The original release paper of Tacotron2 reports a MOS of 4.53, but the model scores much lower on \href{https://paperswithcode.com/sota/text-to-speech-synthesis-on-ljspeech}{LJSpeech}.} 

\textbf{Speech-T5} \cite{ao2022speecht5} is a Transformer-based encoder-decoder. The encoder embeds token indices based on which the decoder predicts a log Mel-filterbank. A HiFi-GAN vocoder \cite{DBLP:journals/corr/abs-2010-05646} is used to convert the predicted log Mel-filterbank to a waveform. The encoder-decoder is jointly pre-trained on speech and text from audiobooks (960h of spoken language and 400M written sentences from LibriSpeech, \citet{panayotov2015librispeech}). For TTS, the model is fine-tuned on 460 hours from LibriTTS \cite{zen2019libritts}. MOS: 3.65

\textbf{Parler-TTS} \cite{lyth2024natural} is a decoder-only Transformer. The model autoregressively predicts latent audio tokens given a sequence of pre-pended text tokens. These audio tokens are then decoded into a waveform using the Descript Audio Codec (DAC) \cite{kumar2023highfidelityaudiocompressionimproved}. We use Parler-TTS Mini v0.1, which was trained on 10k hours from the English portion of Multilingual LibriSpeech \cite{pratap2020mls} plus 585 hours from LibriTTS-R \cite{koizumi2023librittsrrestoredmultispeakertexttospeech}. MOS: 3.92


\section{Experiment 1: Ambiguous Structures}

The goal of this experiment is to assess whether TTS systems can correctly analyze the structure of sentences with (temporary) syntactic ambiguity, and place intonational boundaries in the correct positions accordingly. Using controlled stimuli, we analyze which cues are used by the systems to disambiguate these sentences. 


\subsection{Syntactic Disambiguation}

Garden path sentences contain temporary syntactic ambiguity because the syntactic closure point can either appear early or late in the sentence. Consider the following examples:

\begin{enumerate} \item \textbf{Early closure:} \textit{When Roger left$_\text{A}$ the house was dark.}
\item \textbf{Late closure:} \textit{When Roger left the house$_\text{B}$ it was dark.} \end{enumerate}

In the early closure condition, the syntactic boundary occurs at position \textit{A}, while in the late closure condition, the boundary appears later, at position \textit{B}. The word \textit{was} or \textit{it} resolves the ambiguity. We investigate if TTS systems are sensitive to these syntactic cues and place intonational boundaries in the correct positions accordingly. 

As a control, we use the same sentences with a comma inserted at the syntactic closure point (i.e., \textit{When Roger left,$_\text{A}$ the house was dark} and \textit{When Roger left the house,$_\text{B}$ it was dark}). These commas should provide the systems with more explicit, surface-level cues for generating intonational boundaries. Having this control condition allows us to observe a clear "ground-truth" intonation pattern for each underlying structure.

For our stimuli, we used 45 garden path sentences from several psycholinguistic studies \cite{KJELGAARD1999153,10.1162/jocn.2011.21610}, which are listed in Appendix \autoref{tab:garden-path-stimuli}.

\subsection{Semantic Disambiguation}

In addition to syntactic cues, semantic information can also be used to resolve syntactic ambiguity. To test whether TTS systems are sensitive to semantic cues, we used sentences with attachment ambiguity containing a semantic bias towards either high (VP) or low (NP) attachment. For example:

\begin{enumerate}
\item \textbf{High attachment:} \textit{The boy looked at the painting$_\text{A}$ with much enthusiasm.} 
\item \textbf{Low attachment:} \textit{The boy looked at the painting with muted colours.$_\text{B}$} 
\end{enumerate}

The prepositional phrase \textit{with enthusiasm} is more likely to attach to \textit{looked at}, whereas \textit{with muted colours} is semantically more likely to attach to \textit{the painting}. We analyze if TTS system can distinguish between these structures based on this semantic bias. If so, we would expect an intonational boundary at position \textit{A} to signal the high attachment structure, and no boundary at that position to signal the low attachment structure. Again, we add a control condition with a comma placed at the boundary position, but only for the high attachment cases (e.g., \textit{The boy looked at the painting,$_\text{A}$ with much enthusiasm}), since the comma would be unnatural in the low attachment cases (e.g., \textit{The boy looked at the painting, with muted colours$_\text{B}$}).

\begin{figure*}[ht]
    \centering
    \includegraphics[width=0.9\linewidth]{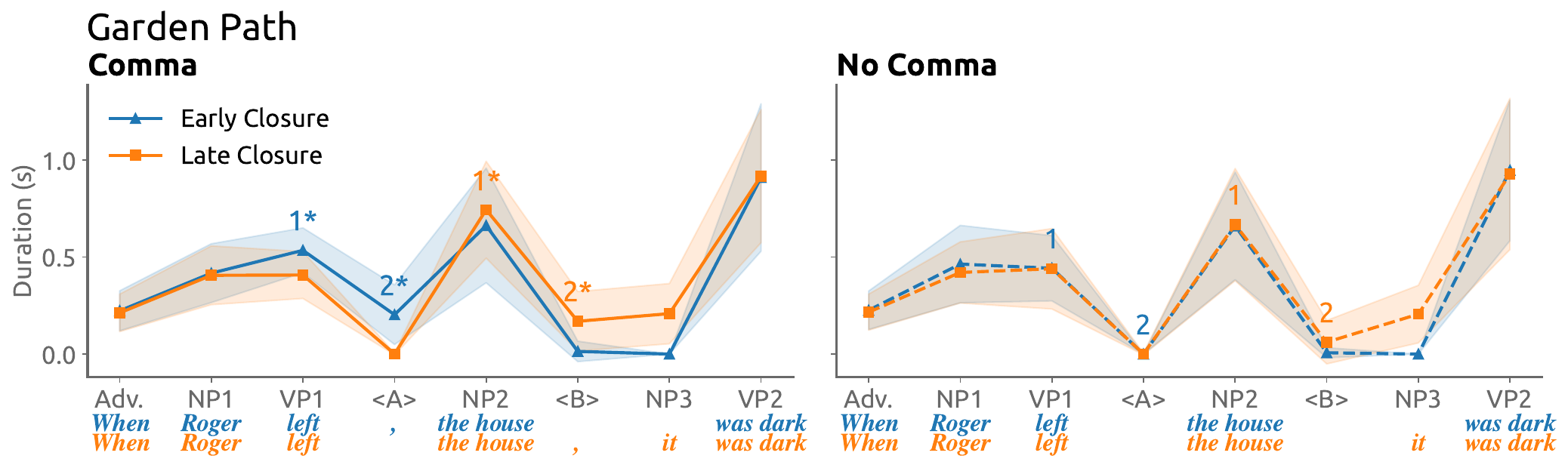}
    \includegraphics[width=0.9\linewidth]{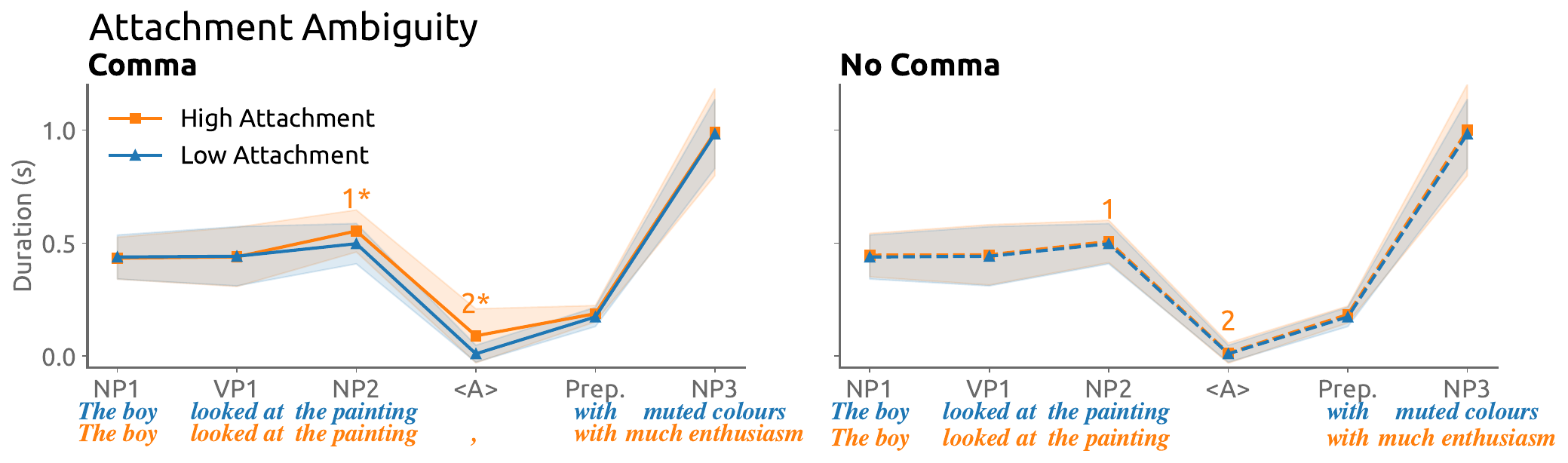}
\caption{Average durations of sentence regions in garden path sentences (top) and sentences with attachment ambiguity (bottom), generated by Parler-TTS. An intonational boundary consists of \textit{lengthening} at the pre-boundary position (1), and insertion of a \textit{pause} at the syntactic boundary position (2); asterisks indicate the presence of these effects. Example sentences are annotated on the x-axes; shading indicates the standard deviation across sentences.}
\label{fig:garden-path}
\end{figure*}

We generated stimuli using the following template: \texttt{<Animate Subject> <Verb> <Inanimate Object> with <Inanimate/Animate Property>}. We filled each slot with six different phrases and generated all possible combinations, resulting in a dataset of 1296 sentences with a semantic bias towards low attachment and 1296 with a bias towards high attachment. Examples are listed in Appendix  \autoref{tab:attach-ambig-stimuli}.



\subsection{Measuring Intonational Boundaries}

We use the Montreal Forced Aligner (MFA) \cite{mcauliffe17_interspeech} to align the generated speech with the input text and measure the duration of each sentence region within the garden path and attachment ambiguity sentences. We identify the presence of intonational boundaries by examining two durational cues: 1) lengthening at the pre-boundary position and 2) the insertion of a pause (i.e., silence, indicated by an unannotated segment by the MFA) at the boundary position. However, we acknowledge that this method has limitations (see \autoref{app:limitations}), as other prosodic cues such as pitch and intensity also contribute to the perception of intonational boundaries.

\subsection{Results}

\autoref{fig:garden-path} shows average durations across sentence regions as generated by Parler-TTS. The results for Tacotron2 and Speech-T5 are highly similar and shown in Appendix \autoref{fig:garden-path-2}.

We observe a strong dependence on comma cues: the systems lengthen the pre-boundary position (1) and insert a pause at the syntactic boundary position (2) only in the presence of a comma at position (2). Without comma cues, the systems default to the statistically most likely intonation pattern. For garden path sentences, this means that no intonational boundaries are generated at position <A>, and occasionally, a pause is inserted at position <B>, since late closure sentences are statistically more likely than early closure sentences. For attachment ambiguity, this means that no intonational boundaries are generated, even if it does not align with the semantic bias of the prepositional phrase.

\section{Experiment 2: Simple Structures}



Our previous experiment indicates that TTS systems struggle to resolve local or global ambiguities in syntactic structure, and are much more dependent on explicit punctuation cues for the generation of intonational boundaries at the correct positions. This is in a sense a human-like effect, as the syntactic structure of garden path and attachment ambiguity sentences is hard to parse, even for humans. It is possible that models correctly incorporate syntactic cues when these are more reliable (i.e., not ambiguous).

In the next experiment, we analyze the role of syntactic cues for intonational boundary placement in simpler sentence structures. We also investigate the role of commas in more detail: are they purely mechanical markers that always trigger a pause, or can TTS systems combine evidence from commas and syntax? To address this, we place commas in syntactically natural and unnatural positions (i.e., aligned with a clause boundary or not), and then compare the strength of the intonational boundaries generated at these points.

\begin{figure*}
    \centering
    \includegraphics[width=1\linewidth]{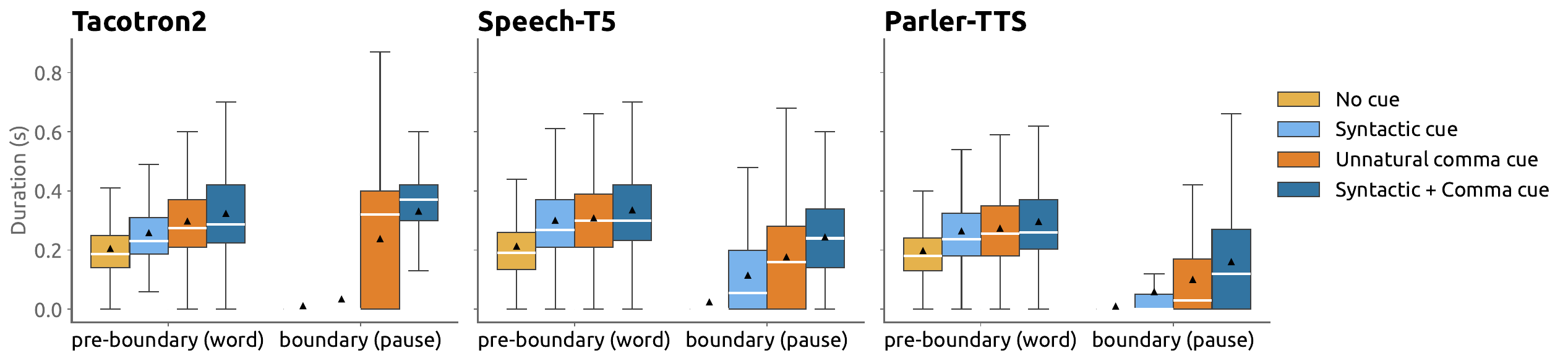}
    \caption{Durations of critical regions (i.e., pre-boundary word and pause at the boundary position), as generated by three TTS systems given different cues: presence or absence of a comma (light vs. dark); measurement of the pause at a syntactic or non-syntactic boundary (blue vs. orange). Black triangles are means, white lines are medians.}
\label{fig:simple-sents}
\end{figure*}

\subsection{Data}

From Simple Wikipedia\footnote{https://simple.wikipedia.org}, we select sentences that contain exactly one comma, marking a syntactic boundary.\footnote{Additional filters were applied: sentences had to be between 7 and 15 words long and free of digits, punctuation (except commas and final periods), and bracketed phrases.}  We select boundaries that signal major structural breaks, which typically lead to an audible intonational boundary in spoken language. We use dependency parsing to detect such structural breaks (examples are listed in Appendix \autoref{tab:syntactic_boundaries}). We create different versions of each sentence, such that the TTS systems have access to different cues for potentially generating an intonational boundary. After synthesizing these sentences, we measure the presence of an intonational boundary at position \textit{A}:

\begin{enumerate}
    \item \textbf{Comma cue + Syntactic cue}: \textit{Most links are blue,$_\text{A}$ but they can be any color}.
    \item \textbf{Syntactic cue}: \textit{Most links are blue$_\text{A}$  but they can be any color}.
\end{enumerate}

In (1), the systems can use the comma at position \textit{A} as a cue for generating an intonational boundary. Additionally, they can use the fact that position \textit{A} is a clause boundary. In (2), the systems can only rely on the clause boundary information, since the comma is absent.

To investigate the extent to which TTS systems generate intonational boundaries at syntactically unnatural positions, we measure the presence of an intonational boundary at position \textit{B}:

\begin{enumerate}
    \setcounter{enumi}{2}
    \item \textbf{Unnatural comma cue}: \textit{Most links are blue but they can,$_\text{B}$  be any color}.
    \item \textbf{No cue}: \textit{Most links are blue but they can$_\text{B}$ be any color}.
\end{enumerate}

In (3), the systems can use the comma as a cue for generating an intonational boundary at position \textit{B} (although it appears at a syntactically unnatural position). In (4), there is no cue that indicates the need for an intonational boundary at position \textit{B}.

\subsection{Evaluation}

Besides comparing the durations for critical regions (i.e., the (pre-)boundary position) across conditions\footnote{The words preceding the syntactic boundary position \textit{A} and non-boundary position \textit{B} may have different lengths, which could affect the average duration. To account for this, we averaged word duration by syllable count.}, we compute  a \textbf{Syntactic Sensitivity Score} for each system. This consists of precision, recall and F1 scores based on the following counts in the sentences without commas: \textbf{True Positives} occur when the model generates a pause at a syntactic boundary (position \textit{A}), \textbf{False Positives} when it generates a pause at a syntactically unnatural position (position \textit{B}), \textbf{False Negatives} when no pause is generated at position \textit{A}, and \textbf{True Negatives} when no pause is generated at position \textit{B}.




\subsection{Results}

\autoref{fig:simple-sents} shows the durations for the pre-boundary word and boundary pause, depending on condition. We see that all models show a similar pattern: the strongest intonational boundaries are produced in the Syntactic + Comma cue condition. None of the models produce an intonational boundary in the No cue condition. The Syntactic cue and Unnatural comma cue conditions are inbetween, with the comma cue leading to a slightly stronger intonational boundary than the syntactic cue. This indicates that in simple sentence structures, TTS systems do pick up on syntactic cues, but that commas simply provide more direct evidence for intonational boundaries. It also shows that models integrate evidence from multiple sources: the combination of a comma and a syntactic cue leads to a stronger intonational boundary than only one cue.

\begin{figure}
    \centering
    \includegraphics[width=0.99\linewidth]{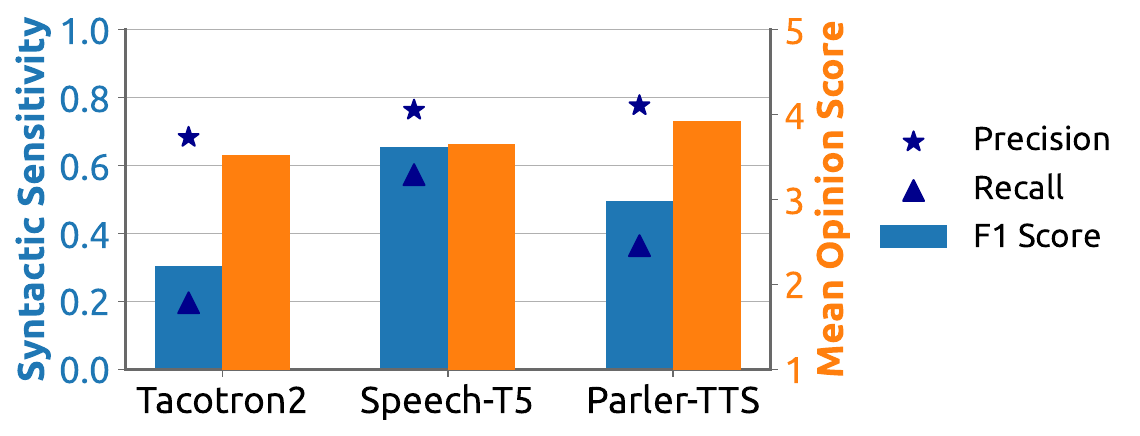}
\caption{Syntactic Sensitivity versus Mean Opinion Score across TTS systems. The F1 score represents the harmonic mean of a system's precision and recall in generating pauses at syntactic boundaries.}
\label{fig:syn-vs-mos}
\end{figure}

In \autoref{fig:syn-vs-mos}, we compare our Syntactic Sensitivity score with reported MOS for each system. We see that \textit{precision} mirrors the MOS pattern (Tacotron2 < Speech-T5 < Parler-TTS), while Speech-T5 has better \textit{recall} than Parler-TTS. In other words: False Positives (i.e., pauses placed at syntactically unnatural positions) seem to affect human ratings more than False Negatives (i.e., no pauses at syntactic boundaries). This illustrates that our Syntactic Sensitivity score provides complementary insights that MOS does not capture. 

\section{Interpreting Boundary Placement}

In the previous experiments, we used controlled stimuli to analyze how two specific cues influence intonational boundary placement in TTS systems. It could be the case, however, that systems' predictions are modulated by the presence of lexical items associated with pauses (e.g., conjunctions such as \textit{but}, \textit{and}, \textit{or}). To gain insight into these cues, we train regression models with a range of different predictor variables to approximate the intonational boundary placement behavior of each TTS system. 


For each TTS system, we train two regression models to predict the following outcome variables for a given position in a sentence: the duration of a pause in that position (\textbf{pause duration}), and the duration of the word before that pause (\textbf{pre-boundary word duration}) (i.e., the two aspects of an intonational boundary we focus on). We again use the sentences from Simple Wikipedia as input and extract the features listed in \autoref{tab:predictors} at the positions marked as \textit{A} or \textit{B} (see Section 5.1).

\paragraph{Implementation} Since we have a large number of (correlated) features, we use LASSO (Least Absolute Shrinkage and Selection Operator; \citealt{tibshirani1996regression}). This regularization technique introduces a penalty term that encourages sparsity in the model, allowing only a subset of features to be used in predicting the outcome variable, preventing overfitting and reducing the effect of multicollinearity (when features are highly correlated, LASSO tends to select only one of them). We apply standardization to the numerical features to ensure they have a mean of zero and a standard deviation of one (unit variance). We train the regression models on 80 percent of the data and evaluate on the remaining 20 percent. We use \( R^2 \) (explained variance) as our evaluation metric to gauge how well the predicted regression lines fit the data.



\begin{table}[ht]
    \centering
    \small
    \begin{tabular}{ll}
        \toprule
        \textbf{Category} & \textbf{Predictor} \\
        \midrule
        \textbf{Punctuation} & Comma Presence (1 or 0) \\
        \midrule
        \textbf{Lexical} & Preceding POS tag (one-hot) \\
        & Following POS tag (one-hot) \\
        \midrule
        \textbf{Constituency} & Is Clause Boundary (1 or 0) \\
        & Num. Closing Brackets \\
        & Max. Tree Depth \\
        \textbf{Dependency} & Preceding Token: Is Dep. Head (1 or 0) \\
        & Preceding Token: Num. Dependents \\
        & Preceding Token: Depth in Subtree \\

        
        \midrule
        \textbf{Length} & Preceding Token Length \\
        & Following Token Length \\
        & Sentence Length \\
        & Number of Preceding Tokens \\
        \midrule
        \textbf{Interaction} & Is Clause Boundary * Comma Presence \\
        \bottomrule
    \end{tabular}
    \caption{Predictor variables for regression models. Global features are extracted from the entire sentence; the other features are extracted at the boundary positions described in Section 5.1.}
    \label{tab:predictors}
\end{table}

\subsection{Results}

\begin{table}[ht]
\centering
\small
\begin{tabular}{lcc}
\toprule
\textbf{Model} & \textbf{Pause Dur.} & \textbf{Pre-boundary Word Dur.} \\
\midrule
Parler-TTS & .14 & .37 \\
Speech-T5  & .30 & .44\\
Tacotron2  & .44 & .42 \\
\bottomrule
\end{tabular}
\caption{Explained variance (\( R^2 \)) of linear regression models for predicting \textit{pause duration} and \textit{pre-boundary word duration} as generated by three different TTS systems. Reported scores are for a held-out test set.}
\label{tab:tts_metrics}
\end{table}

\paragraph{Performance} The performance of the regression models is displayed in \autoref{tab:tts_metrics}. We see that our predictor variables generally explain more variance in the \textit{pre-boundary word duration} data compared to the \textit{pause duration} data, which makes sense given that we use explicit features of the pre-boundary word (e.g., its length). We also see that pause duration is more predictable for Tacotron2 than for the other two systems. The behavior of Parler-TTS is least predictable, indicating that this model relies on other features than the ones we included in our regression models, or on more complex interactions between those features.

\paragraph{Feature Importance}

\begin{figure}
    \centering
    \includegraphics[width=0.96\linewidth]{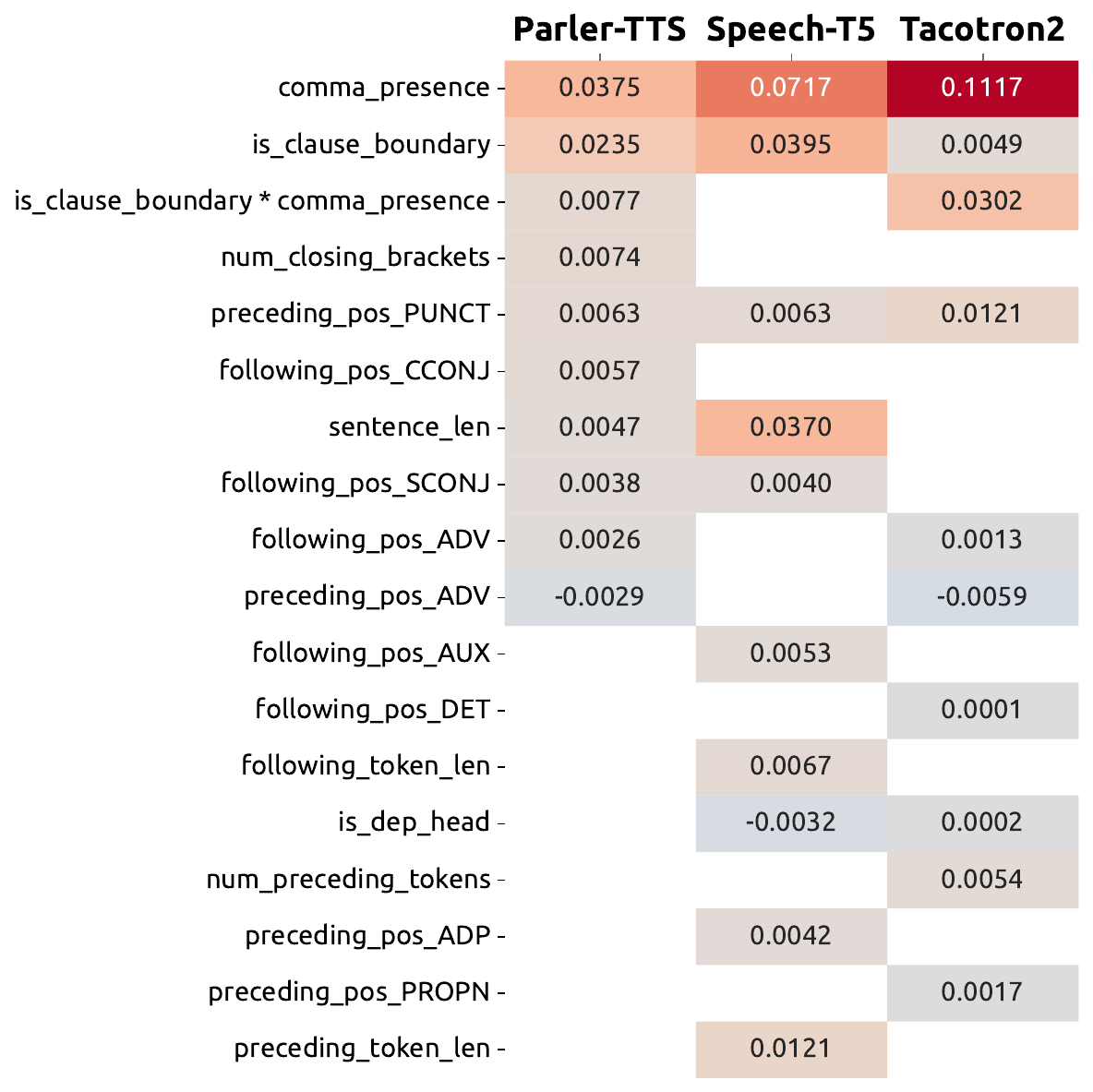}
\caption{Coefficients of LASSO-selected predictor variables for pause durations of TTS systems.}
\label{fig:feat-importance}
\end{figure}

\autoref{fig:feat-importance} shows the top 10 selected predictors for \textit{pause duration} for each of the TTS systems, together with their regression coefficients. We see that \textit{comma presence} is the strongest predictor for all three TTS systems, verifying their strong reliance on punctuation cues. For Parler-TTS and Speech-T5, \textit{is clause boundary} is also an important predictor.\footnote{We verified that \textit{is clause boundary} was a predictor by itself by running LASSO on different subsets of sentences: with/without commas, and with/without predictive lexical items (e.g., conjunctions). In all cases, \textit{is clause boundary} was still selected as an important predictor.} We also see that specific lexical items are selected, e.g., words with the POS tag SCONJ or CCONJ. Depending on the model, different length-related features are also selected: \textit{sentence length} for Parler-TTS and Speech-T5, \textit{preceding/following token length } for Speech-T5, and \textit{num. preceding tokens} for Tacotron2.


Overall, the analysis confirms that punctuation plays a major role in determining the duration of intonational boundaries in TTS systems. It also demonstrates that specific lexical items and length-related features influence pause duration. This reliance on surface cues is particularly evident in the LSTM-based system Tacotron2, while the Transformer-based systems Parler-TTS and Speech-T5 also seem to incorporate some syntactic information.

\section{Changing the Training Distribution}


While TTS systems may see plenty of examples of simple syntactic structures with obvious clause boundaries, garden path sentences are likely underrepresented in their training data. Sentences with attachment ambiguity may occur more frequently. However, even for such sentences, the intonation patterns we aim to capture (where high attachment introduces an intonational boundary and low attachment does not) may still be rare in the training data. As discussed in Section 2, speakers use distinct intonation patterns to disambiguate high and low attachment in conversational settings, helping to convey the intended meaning. In non-conversational speech, this distinction is less frequently observed. Consequently, TTS systems trained on audiobooks may not have sufficient exposure to the nuanced intonation patterns associated with the different syntactic structures.

\subsection{Training data analysis}
Out of the three TTS systems we investigated, Parler-TTS was trained on the largest amount of data. To check if it missed important evidence for high and low attachment structures, we selected a subset of the MLS corpus that Parler-TTS was trained on (5000 examples, \textasciitilde12000 sentences) and counted the occurrences of pauses, commas, and frequent prepositions\footnote{\textit{of, to, in, for, with, as, at, on, by, for}.}, as well as the overlap between them. The detailed results are shown in Appendix \autoref{fig:venn-diagram}. While we cannot directly determine how often the model encountered high or low attachment structures, we observe that prepositions \textit{without} a preceding pause (aligning with low attachment) appeared almost 5 times more frequently than those \textit{with} a preceding pause (aligning with high attachment). This imbalance may explain why the model struggles to generate distinct intonation patterns for the two structures.

\subsection{Altering the training distribution}
We hypothesize that a greater balance in the occurrence of high and low attachment structures in the training data will enable the model to generate more varied intonation patterns that better reflect the underlying structure. To test this hypothesis, we conducted two finetuning experiments aimed at rebalancing the data. These experiments are not meant to directly improve the performance of Parler-TTS, but merely to diagnose the role of (lack of) exposure to certain structures.

\paragraph{Finetuning on sampled data} For the first experiment, we selected all sentences from the Jenny corpus\footnote{\url{https://github.com/dioco-group/jenny-tts-dataset}} containing a preposition preceded by a pause (\textasciitilde5000 sentences, \textasciitilde6 hours of speech). To ensure that the model would not be able to rely on commas as a cue for generating intonational boundaries, we removed all commas from the transcriptions. Our hope was that showing the model more examples of \textit{general} PPs preceded by a pause would lead to more varied intonation patterns for sentences with an \textit{ambiguous} PP.

\paragraph{Finetuning on synthetic data} For the second experiment, we created a synthetic dataset to provide the model with more explicit examples of high and low attachment. Using the template described in Section 4.2, we generated 2500 sentences with a semantic bias towards high attachment, and 2500 sentences with a bias towards low attachment (resulting in \textasciitilde6 hours of speech). We synthesized these sentences using Tacotron2, inserting commas at positions that would correspond to intended pauses (e.g., before the preposition \textit{with} in high attachment cases). We again removed these commas from the text to ensure that the model could not rely on punctuation, but instead learn to use the semantic bias of the sentences to predict the presence of a pause.

\paragraph{Evaluation} We created an evaluation set consisting of sentences containing function words that could be interpreted in two different ways, with one interpretation requiring a pause before the word (e.g., \textit{The boy looked at the painting <pause> with genuine interest}) and the other not (e.g., \textit{The boy looked at the painting with muted colors}). These function words include \textit{with} (our primary example for high and low attachment), but also \textit{as}, \textit{for}, and \textit{to}, as shown in \autoref{tab:prep-examples} in the Appendix. We created 30 sentences per category and sampled them three times from the models (using three different random seeds). We then measured the pause duration before the critical function word across the resulting 90 data points.


\subsection{Results}

\begin{figure}
    \centering
    \includegraphics[width=0.72\linewidth]{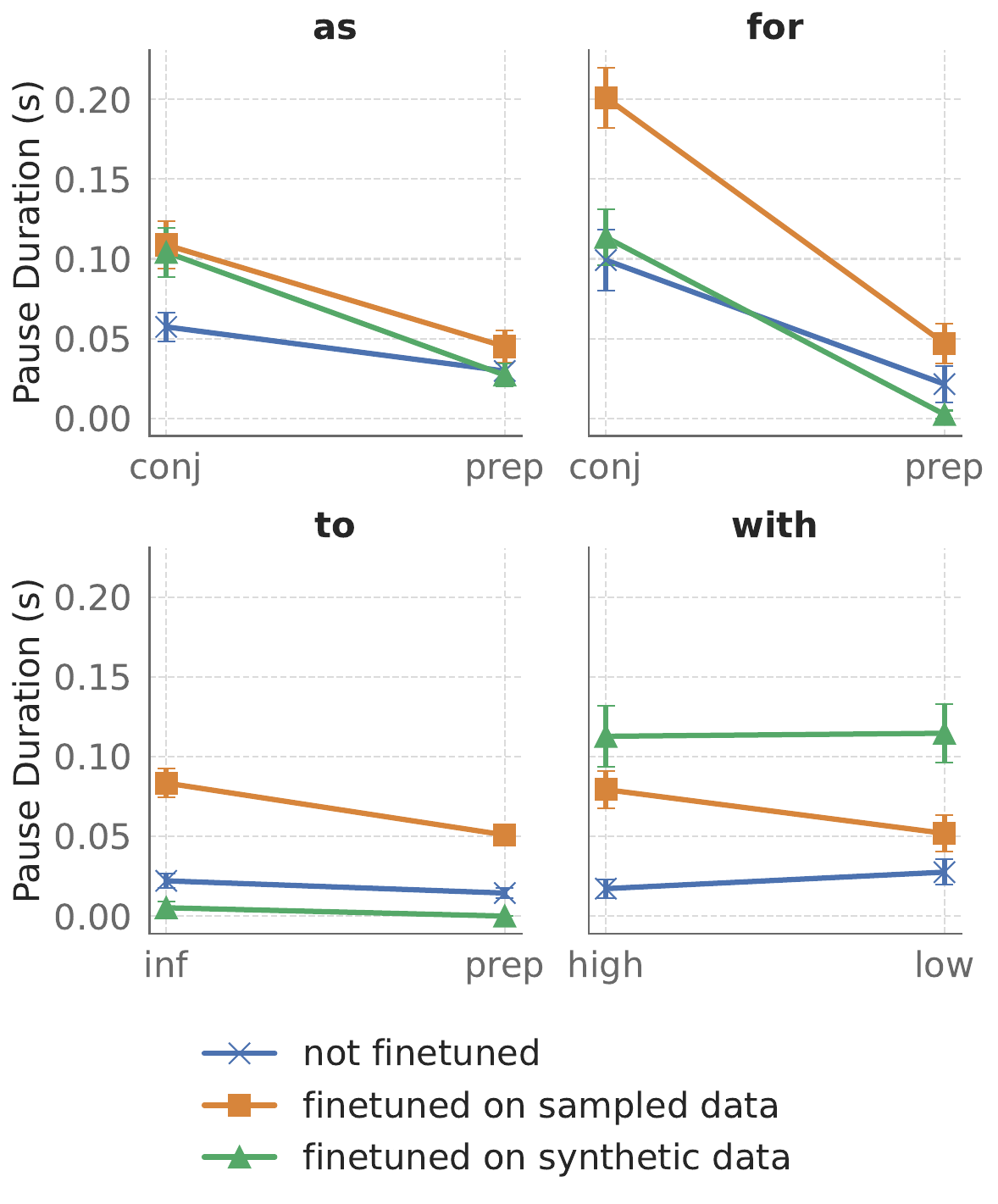}
\caption{Average pause duration before the function words \textit{as, for, to, with} (each used in two different ways, e.g., as a preposition vs. conjunction) for three versions of Parler-TTS. Error bars indicate the standard error.}
\label{fig:finetuning-experiments}
\end{figure}


\autoref{fig:finetuning-experiments} shows that the model finetuned on sampled data (orange lines) generates longer pauses than the non-finetuned model (blue lines). Interestingly, this increase is more pronounced in contexts where a pause is expected, i.e., before \textit{for} and \textit{as} when used as a conjunction, before \textit{to} when used as an infinitive, and before \textit{with} in the high attachment case. This suggests that training the model on a more balanced data distribution leads to more distinct intonation patterns that reflect different syntactic structures. 

In contrast, the model finetuned on synthetic data did not learn to distinguish between high and low attachment based on semantic cues, as the pause duration before \textit{with} remains the same in both cases (although it did increase compared to the non-finetuned model). These results indicate that, even with more exposure, TTS systems cannot disambiguate syntactic structure based on semantic cues. However, this observation requires further investigation, particularly regarding the role of natural versus synthetic speech and the amount of data necessary for robust results.

\section{Conclusion}
We evaluated the syntactic sensitivity of TTS systems by analyzing their intonation patterns generated for controlled stimuli. We find that systems can identify obvious clause boundaries in simple sentences but struggle with more complex, locally or globally ambiguous structures. We also investigated the role of (lack of) exposure to such structures, and show that systems can generate more syntax-aligned intonation patterns if provided with appropriate evidence.



Future work should study a broader range of phenomena to better understand the types of linguistic associations captured by TTS systems. One potential direction would be to develop a resource similar to BLiMP \cite{warstadt-etal-2020-blimp-benchmark} for TTS, which could serve as a more comprehensive framework for evaluating their syntactic sensitivity. Additionally, \textit{structural probing} \cite{hewitt2019structural,shen23_interspeech} could offer a more detailed look at the internal representation of syntax in TTS systems, complementing our behavioral measures.

\section{Acknowledgements}
We are grateful to Grzegorz Chrupała, Jaap Jumelet and Tom Lentz for reviewing early versions of this paper. This research is funded by the Netherlands Organisation for Scientific Research (NWO) through NWA-ORC grant NWA.1292.19.399 for \textit{InDeep}.

\bibstyle{acl_natbib}
\bibliography{anthology,custom}

\newpage
\appendix
\section{Appendix Figures and Tables}
\begin{figure*}[ht]
    \centering
    \includegraphics[width=1\linewidth]{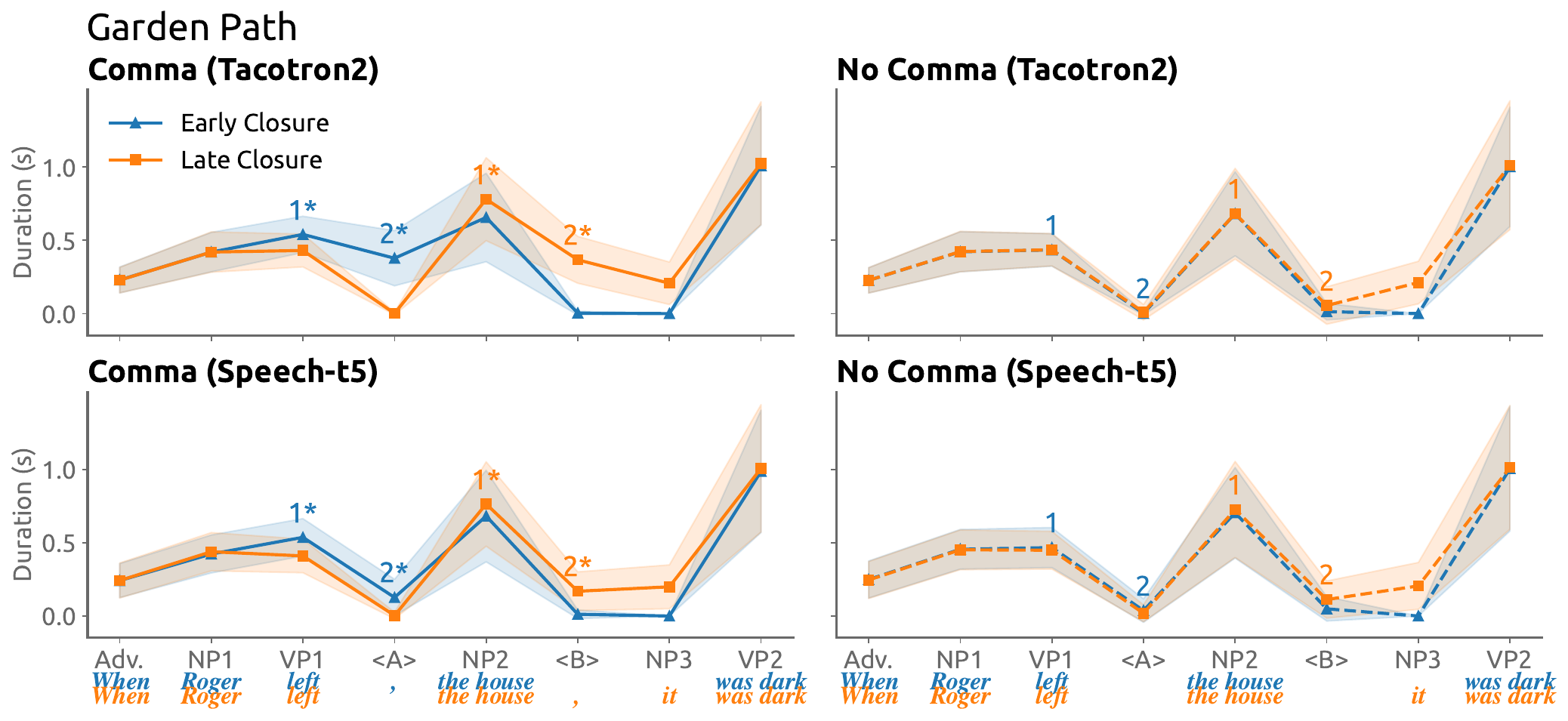}
    \includegraphics[width=1\linewidth]{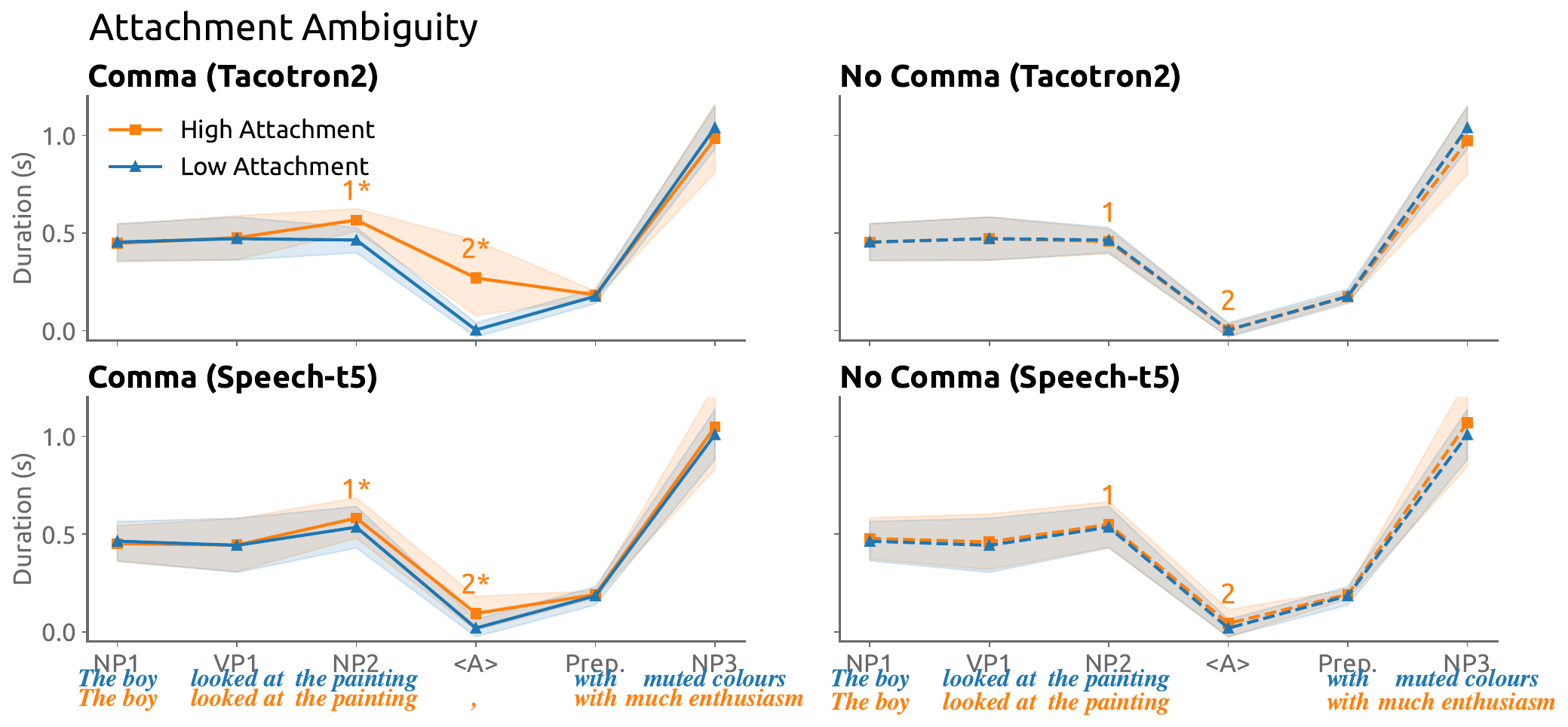}
\caption{Average durations of sentence regions in garden path sentences (top) and sentences with attachment ambiguity (bottom), generated by Tacotron2 and Speech-T5. An intonational boundary consists of \textit{lengthening} at the pre-boundary position (1), and insertion of a \textit{pause} at the syntactic boundary position (2); asterisks indicate the presence of these effects. Example sentences are annotated on the x-axes; shading indicates the standard deviation across sentences.}
\label{fig:garden-path-2}
\end{figure*}

\begin{figure}[ht]
    \centering
    \includegraphics[width=1\linewidth]{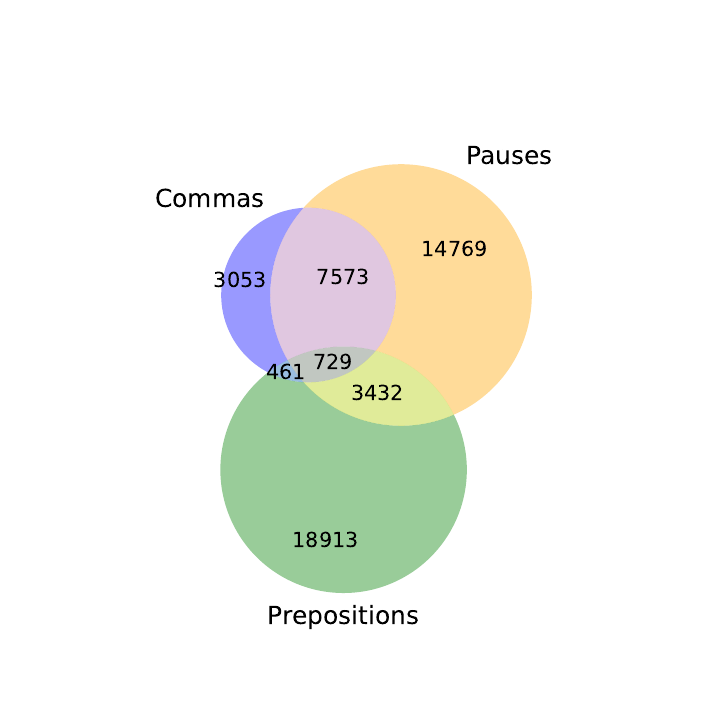}
\caption{Counts of frequent prepositions, commas and pauses, as well as their overlap, in a subset of the training data of Parler-TTS.}
\label{fig:venn-diagram}
\end{figure}

\autoref{fig:garden-path-2}, \autoref{fig:venn-diagram} and \autoref{tab:garden-path-stimuli}, \autoref{tab:attach-ambig-stimuli}, \autoref{tab:syntactic_boundaries}, \autoref{tab:prep-examples} are shown on the next pages.

\section{Limitations}\label{app:limitations}
This study has several limitations that should be acknowledged. First, we analyzed intonational boundaries based on duration measures only. While pause duration and word lengthening are well-established proxies for intonational boundaries, other prosodic features (e.g., pitch contour and intensity) also contribute significantly to their perception. Although previous research suggests that duration measures alone can reliably indicate the presence of an intonational boundary, and that pitch and intensity are less consistent across speakers and contexts \cite{KJELGAARD1999153,10.1162/jocn.2009.21269,10.1162/jocn.2011.21610}, incorporating these additional prosodic cues would allow us to better characterize intonational structure (as generated by TTS systems).

Second, we did not  consider different levels of boundary strength, a distinction made by the Tones and Break Indices (ToBI) framework \cite{silverman1992tobi}. Future work could benefit from adopting this gradation to more fully capture the complexity of intonational phrasing.

Third, the Parler-TTS model supports conditioning on voice characteristics specified through natural language descriptions. However, in this study, we only used a single voice description to synthesize our stimuli. It remains an open question how varying these voice characteristics might influence the resulting intonation patterns.

\begin{table*}[ht]
\centering
\small
\caption{Garden path stimuli for Experiment 1. Sentences were presented in two forms: \textbf{early closure} (without the word in brackets) and \textbf{late closure} (with the word in brackets).}
\begin{tabular}{l}
\toprule
\textbf{Stimulus} \\
\midrule
Whenever John walks the dogs (\textit{cats}) are chasing him. \\
Because John studied the material (\textit{it}) is clearer now. \\
When Whitesnake plays the music (\textit{it}) is loud. \\
When Tim presents the lectures (\textit{they}) are interesting. \\
When the original cast performs the plays (\textit{they}) are funny. \\
When Madonna sings the song (\textit{it}) is a hit. \\
Whenever John swims the channel (\textit{it}) is choppy. \\
When Roger left the house (\textit{it}) was dark. \\
Whenever Frank performs the show (\textit{it}) is fantastic. \\
Because Mike phoned his mother (\textit{she}) is relieved. \\
When the clock strikes the hour (\textit{it}) is midnight. \\
Whenever the guard checks the door (\textit{it}) is locked. \\
If Laura folds the towels (\textit{they}) are neat. \\
If George programs the computer (\textit{it}) is sure to crash. \\
If Charles babysits the children (\textit{they}) are happy. \\
When the maid cleans the rooms (\textit{they}) are immaculate. \\
Before Jack deals the cards (\textit{they}) are shuffled. \\
While the boy read books (\textit{televisions}) were stolen. \\
When the dog bites cats (\textit{mice}) run away. \\
When the man batted balls (\textit{players}) covered the field. \\
While the man parked cars (\textit{bikes}) were waiting. \\
After the puppy licked kids (\textit{parents}) were laughing. \\
Because snakes eat mice (\textit{toads}) hide. \\
When a bear approaches people (\textit{dogs}) come running. \\
After the chef cooked cake (\textit{coffee}) was served. \\
While the artist painted clouds (\textit{stars}) were appearing. \\
As the cat climbed trees (\textit{leaves}) were falling. \\
As John hunted the frightened deer (\textit{it}) escaped through the woods. \\
When Anne visited the British relatives (\textit{they}) were moving to London. \\
When Rita washed her favorite sweater (\textit{it}) was torn to shreds. \\
When Joan left her old boyfriend (\textit{he}) stalked her for two months. \\
While the assistant measured the delicate fabric (\textit{it}) ripped and frayed. \\
When Greg returned the new car (\textit{it}) was operating smoothly. \\
Because Cecelia baked the delicious homemade bread (\textit{it}) was served at breakfast. \\
Even when Todd cleaned the small kitchen (\textit{it}) smelled like old garbage. \\
Because Grandma knitted wool sweaters (\textit{they}) would appear under the Christmas tree. \\
Because Maria read the financial news (\textit{it}) was always at her fingertips. \\
As Sam pounded the thin metal (\textit{it}) ripped and broke into pieces. \\
When Sonya painted the kitchen walls (\textit{they}) were covered into obvious drops. \\
As Lia typed the eviction notice (\textit{it}) was cancelled. \\
When Tina supervised the night crew (\textit{it}) was very efficient. \\
As Gary watched the drunken workmen (\textit{they}) stumbled off the platform. \\
When the sheriff patrolled the whole area (\textit{it}) was very safe. \\
When the musician conducted the symphony orchestra (\textit{it}) was at its peak. \\
When Molly sang the drinking songs (\textit{they}) sounded like opera. \\
\bottomrule
\end{tabular}
\label{tab:garden-path-stimuli}
\end{table*}

\begin{table*}
\centering
\small
\caption{Examples of attachment ambiguity stimuli for Experiment 1. Two prepositional phrases were constructed for each stimulus, the former creating a semantic bias towards high (VP) attachment, the latter creating a semantic bias towards low (NP) attachment.}
\begin{tabular}{l}
\toprule
\textbf{Stimulus} \\
\midrule
The boy looked at the painting \textit{with much enthusiasm} / \textit{with muted colors}. \\
The woman described the table \textit{with much enthusiasm} / \textit{with the smooth surface}. \\
The man bought the vase \textit{with much happiness} / \textit{with red dots}. \\
The girl found the chair \textit{with much ease} / \textit{with blue stripes}. \\
The artist inspected the house \textit{with much interest} / \textit{with wooden details}. \\
\bottomrule
\end{tabular}
\label{tab:attach-ambig-stimuli}
\end{table*}

\begin{table*}[ht]
\centering
\small
\caption{Example sentences and counts of selected dependency labels, taken from the Simple Wikipedia Corpus.}
\begin{tabular}{lll}
\toprule
\textbf{Dependency Label} & \textbf{Example} & \textbf{Count} \\
\midrule
Conjunction (conj) & Most links are blue, but they can be any color. & 420 \\
Adverbial clause modifier (advcl) & Unless the cache is cleared, the link will always stay dark blue. & 161 \\
Relative clause modifier (relcl) & Animals are eukaryotes with many cells, which have no rigid cell walls. & 49 \\
Appositional modifier (appos) & Almost all animals have neurons, a signalling system. & 47 \\
Clausal complement (ccomp) & In Thailand, stingray leather is used in wallets and belts. & 67 \\
Open clausal complement (xcomp) & Genes say to the cell what to do, like a language. & 70 \\
\bottomrule
\end{tabular}
\label{tab:syntactic_boundaries}
\end{table*}

\begin{table*}[ht]
\centering
\small
\begin{tabular}{@{}lll@{}}
\toprule
\textbf{Category}          & \textbf{Pause} & \textbf{Example}                                                                 \\ \midrule
as (preposition)           & no             & She was hired as the new manager of the team.                                     \\
as (conjunction)           & yes            & She left early as she had an important meeting to attend.                        \\
for (preposition)          & no             & The child picked up the toy for his friend who had dropped it.                   \\
for (conjunction)          & yes            & The child picked up the toy for he wanted to play with it.                       \\
to (preposition)           & no             & The man gave the book to his sister who wanted it.                                \\
to (infinitive)            & yes            & The man read the book to learn more about history.                                \\
with (preposition, high attach.)  & yes            & The boy looked at the painting with genuine interest.                             \\
with (preposition, low attach.)   & no             & The boy looked at the painting with muted colors.                                \\ \bottomrule
\end{tabular}
\caption{Example sentences for our evaluation set for the fine-tuning experiments: each function word can be used in two different ways, one of which is associated with a pause.}
\label{tab:prep-examples}
\end{table*}

\end{document}